\documentclass[format=acmsmall, review=false, screen=true]{acmart}

\usepackage{booktabs} 
\usepackage{amsmath}

\usepackage[ruled]{algorithm2e} 

\SetAlFnt{\small}
\SetAlCapFnt{\small}
\SetAlCapNameFnt{\small}
\SetAlCapHSkip{0pt}
\IncMargin{-\parindent}

\acmYear{2019}
\acmMonth{11}
\copyrightyear{2019}

\setcopyright{acmlicensed}

\acmDOI{0000001.0000001}

\begin{document}
\title[Neural Open Relation Tagging Model]{Hybrid Neural Tagging Model for Open Relation Extraction}

\author{Shengbin Jia}
\author{Yang Xiang}
\authornote{This is the corresponding author}

\affiliation{%
  \institution{Tongji University}
  \department{College of Electronic and Information Engineering}
  \city{Shanghai}
  \country{China}}
\email{{shengbinjia,shxiangyang}@tongji.edu.cn}

\thanks{This work is supported by the National Natural Science Foundation of China under Grant No.71571136, and the 2019 Tencent Marketing Solution Rhino-Bird Focused Research Program.}

\begin{abstract}
	
Open relation extraction (ORE) remains a challenge to obtain a semantic representation by discovering arbitrary relation tuples from the unstructured text. Conventional methods heavily depend on feature engineering or syntactic parsing, they are inefficient or error-cascading. Recently, leveraging supervised deep learning structures to address the ORE task is an extraordinarily promising way. However, there are two main challenges: (1) The lack of enough labeled corpus to support supervised training; (2) The exploration of specific neural architecture that adapts to the characteristics of open relation extracting. In this paper, to overcome these difficulties, we build a large-scale, high-quality training corpus in a fully automated way, and design a tagging scheme to assist in transforming the ORE task into a sequence tagging processing. Furthermore, we propose a hybrid neural network model (HNN4ORT) for open relation tagging. The model employs the Ordered Neurons LSTM to encode potential syntactic information for capturing the associations among the arguments and relations. It also emerges a novel Dual Aware Mechanism, including Local-aware Attention and Global-aware Convolution. The dual aware nesses complement each other so that the model can take the sentence-level semantics as a global perspective, and at the same time implement salient local features to achieve sparse annotation. Experimental results on various testing sets show that our model can achieve state-of-the-art performances compared to the conventional methods or other neural models.
	
\end{abstract}

%
%


%
%


\maketitle

\renewcommand{\shortauthors}{S. Jia et al.}

\section{Introduction}

Open relation extraction (ORE)~\cite{Banko2007, Fader2011} is an important NLP task for discovering knowledge from unstructured text. It does not identify pre-defined relation types like the Traditional relation extraction (TRE)~\cite{doddington2004automatic,mintz2009distant,zhu2019graph}. Instead, it aims to obtain a semantic representation that comprises a arbitrary-type relational phrase and two argument phrases. Such as tracking a fierce sports competition, the news reporting works should find unpredictable relation descriptions (e.g., (Gatlin, again won, the Championships). ORE systems can satisfy such personalized scenarios by discovering resourceful relations without pre-defined taxonomies. 
A robust open relation extractor, especially working with limited well-annotated data, can greatly prompt the research not only in knowledge base construction~\cite{carlson2010toward}, but also in a wide variety of applications~\cite{mausam2016open}, for example, intelligent questions and answers~\cite{Fader2014Open,sun2019pullnet}, text comprehension~\cite{Lin2017Reasoning,shah2019open,zhang2019openki}. ORE has gained consistent attention.

However, this task is constrained by the lack of labeled corpus or inefficient extraction models. Conventional methods are usually based on pattern matching to carry out unsupervised or weak-supervised learning. The patterns are usually syntax-relevant. They
are handcrafted by linguists~\cite{Fader2011, DelCorro2013, Angeli2015, jia2018} or bootstrapped depend on syntax analysis tools~\cite{Mausam2012, mausam2016open, bhutani2016nested}. As for the former, the manual cost is high and the scalability is poor. The latter heavily depends on syntax analysis tools, and the cascading effect caused by parsing errors is serious.

Deep learning-based methods are popular and have achieved good accomplishments in various information extraction tasks~\cite{sutskever2014sequence, huang2015bidirectional, Wang2016, zheng2017}. Recently, a few people are trying to solve the ORE task with supervised neural network methods, especially the neural sequence tagging models~\cite{leicui, Stanovsky2018Supervised}. 
Nevertheless, there are many challenges and difficulties. 
(1) There are few public large-scale labeled corpora for training supervised ORE learning models.  And labeling a training set manually with a large number of relations is heavily costly. 
(2) It's always challenging for open-ended tasks to produce supervised systems. The specific neural architecture that adapts to the characteristics of ORE should be explored urgently. 
Normal sequence labeling models encode word-level context information as assigning each word a tag to indicate the boundary of adjacent segments~\cite{Zhuo2016Segment, AAAI1714776}. However, relation extraction is to only mark relational words corresponding to arguments, multiple relations involved in one sentence only focus on fractional aspects of the text. How to get the relevant information among the arguments and a relation,
and how to focus on the key parts of a sentence according to a certain relation, are worthy of special attention.

We overcome the above challenges by the following works. 
As for the first challenge, we build a large-scale, high-quality corpus in a fully automated way. We verify that the dataset has good diversification and can motivate models to achieve promising performances. 
Besides, we design a tagging scheme to transform the extraction task into a sequence tagging problem. A triple corresponds to its unique tag sequence, thus overlapping triples in a sentence can be presented simultaneously and separately.

As for the second challenge, to adapt to the characteristics of open relation extraction, we present a hybrid neural network model (HNN4ORT) for open relation tagging. The model has two highlights: 
(a) It employs the Ordered Neurons LSTM (ON-LSTM) to learn temporal semantics, while capturing potential syntactic information involved in natural language, where syntax is important for relation extraction to acquire the associations between arguments and relations;
(b) We propose the Dual Aware Mechanism, including Local-aware Attention and Global-aware Convolution. The two complement each other and realize the model focusing on salient local features to achieve sparse annotation while  considering a sentence-level semantics as a global perspective.

In summary, the main contributions of this work are listed as follows:
\begin{itemize}
	
	\item We transform the open relation extraction task into a sequence tagging processing and design a tagging scheme to address multiple overlapping relation triples. 
	
	\item We present a hybrid neural network model (HNN4ORT), especially the Dual Aware Mechanism, to adapt to the characteristics of the ORE task.
	
	\item We construct an automatic labeled corpus~\footnote{The dataset can be obtained from https://github.com/TJUNLP/NSL4OIE.} in favor of adopting the supervised approaches for the ORE task. It is simple to construct and is larger-scale than other existing ORE corpora.

	\item Experimental results on various testing sets show that our model can produce state-of-the-art performances compared to the conventional methods or other neural models.
	
\end{itemize}

The rest of this article is organized as follows. In Section 2, we review some related work. Then we detail the process of preparing the training corpus in Section 3. In Section 4, we present our model in detail. In Section 5 and 6, we conduct and analyze experiments on multiple datasets. Section 7, concludes this work and discusses future research directions.

\section{Related Works}
\label{section-two}

Relation extraction can be divided into two branches, including traditional relation extraction and open relation extraction.  

Traditional relation extraction (TRE)~\cite{doddington2004automatic,Culotta2006Integrating,mintz2009distant,zhu2019graph} can be regarded as a classification task which is committed to identifying pre-defined relation taxonomies between two arguments, and undefined relations will not be found. The mainstream methods were based on neural systems to recognize relations by using the large-scale knowledge bases to provide distant supervision~\cite{Xu2015,Wang2016,shi2019brief}.

As for the Open relation extraction (ORE)~\cite{niklaus2018survey,glauber2018systematic}, most of the existing methods used patterns-based 
matching approaches by leveraging linguistic analysis. 
In general, the patterns were generalized by handcrafting~\cite{Angeli2015,DelCorro2013, jia2018} or semi-supervised learning~\cite{Mausam2012, mausam2016open, bhutani2016nested} such as bootstrapping. These manual patterns were higher-accuracy than those produced automatically. However, they were heavy-cost and poor-efficiency.

Many extractors, such as TextRunner~\cite{Banko2007}, WOE$^{pos}$~\cite{Weld2010}, Reverb~\cite{Fader2011}, focused on efficiency by restricting the shallow syntactic parsing to part-of-speech tagging and chunking. Meanwhile, many approaches designed complex patterns from complicated syntactic processing, especially dependency parsers, such as WOE$^{parse}$~\cite{Weld2010}, PATTY~\cite{Nakashole2012}, OLLIE~\cite{Mausam2012}, Open IE-4.x~\cite{mausam2016open}, MinIE~\cite{Gashteovski2017} and so on. These extractors could get significantly better results than the extractors based on shallow syntax. However, they heavily relied on the syntactic parsers. Many papers~\cite{Mausam2012, DelCorro2013} analyzed the errors made by their extractors and found that parser errors account for a large even the largest part of the whole. Parsing errors restrained the extracting performances and would produce a serious error cascade effect.
Taking a certain strategy to refine the extraction results, should be an effective way to gather high-quality relation triples.

The sequence tagging tasks, such as Chinese word segmentation (CWS), Part-of-speech tagging (POS), and Named entity recognition (NER), require to assign representative labels for each word in a sentence. Conventional models were linear statistical models, which included Hidden Markov models~\cite{Baum1966Statistical}, Maximum entropy models~\cite{mccallum2000maximum}, and Conditional random fields (CRF)~\cite{lafferty2001conditional} and so on. Neural methods mapped input sequences to obtain fixed dimensional vector representations by various neural networks~\cite{huang2015bidirectional, collobert2011natural,zhai2017,zhang2019using,guan2019new}, then predicted the target sequences from the vectors using a layer with Softmax activation function~\cite{Chiu2016, huang2015bidirectional} or a special CRF layer~\cite{raganato2017neural}.

There were a few examples that applied neural models to open information extraction tasks.
According to the Machine translation mechanism, the extraction process was converted into text generation. Zhang et al.~\cite{Zhang2017MT} extracted predicate-argument structure phrases by using a sequence to sequence model. Cui et al.~\cite{leicui} proposed a multi-layered encoder-decoder framework to generate relation tuples related sequences with special placeholders as a marker. 
Similarly, Bhutani et al.~\cite{bhutani2019open} used the encoder-decoder method to generate relation triples, but only from the limited question and answer pairs. In addition, Zheng et al.~\cite{zheng2017} creatively designed the model to transform the traditional relation extraction into relation sequence tagging. Later, Stanovsky et. al.~\cite{Stanovsky2018Supervised} formulated the ORE task as a sequence tagging problem. They applied the LSTM with a Softmax layer to tag each word. However, we try to design more effective semantic learning frameworks to annotate relations. Besides, we have a larger training set of about several hundred thousand, but they have only a few thousand.

\section{Training Corpus and Tagging Scheme}
\label{section-three}

There are few public large-scale labeled datasets for the ORE task. Stanovsky and Dagan~\cite{Stanovsky2016Creating} created an evaluation corpus by an automatic translation from QA-SRL annotations~\cite{He2015Question}. It only contains 10,359 tuples over 3200 sentences. Then, Stanovsky et al.~\cite{Stanovsky2018Supervised} further expand 17,163 labeled open tuples from the QAMR corpus~\cite{michael2018qamr}. However, the accurate has declined. 

Therefore, we adopt a mechanism of bootstrapping by using multiple existing open relation extractors without having to resort to expensive annotation efforts. Currently, the training set contains 477,701 triples. And it is easy to get expanded. We also design a tagging scheme to automatically annotate them.

\begin{figure*}
	\centering
	\includegraphics[width=0.8\textwidth]{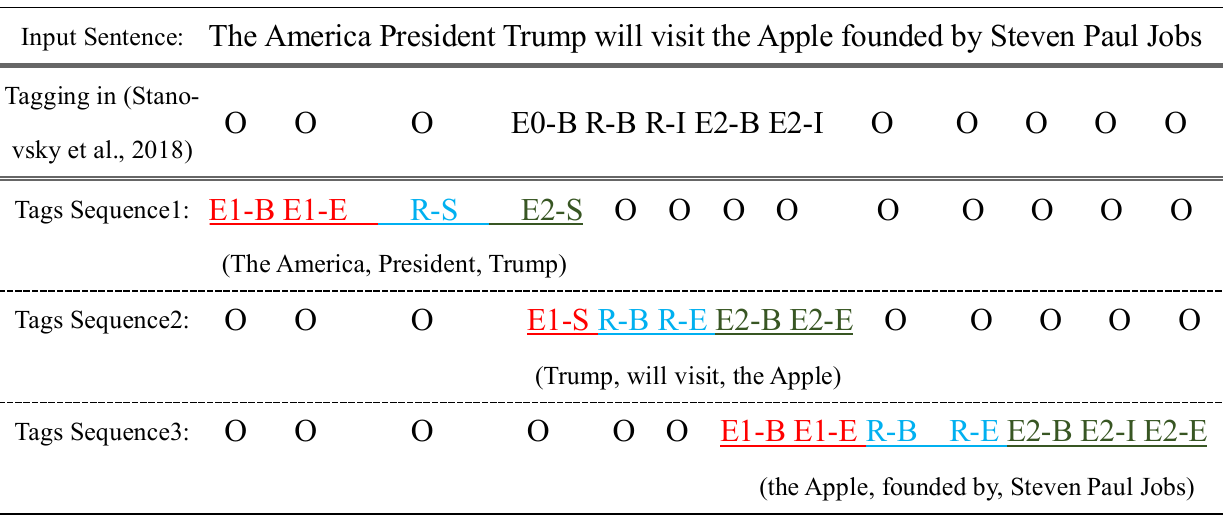}
	\caption{
		Stanovsky et al.~\cite{Stanovsky2018Supervised} annotates the sequence of relations as shown in the 2nd line. A sentence produces only one tags sequence. However, the tagging based on our scheme is shown in lines 3-5. Multiple, overlapping triples can be represented, where ``E1", ``R", ``E2" respectively represent Argument1, Relation, and Argument2. ``Tags Sequence'' is independent of each other.}
	\label{fig:one}
\end{figure*}

\subsection{The Correct Relation Triples Collecting}

We use three existing excellent and popular extractors (OLLIE~\cite{Mausam2012}, ClausIE~\cite{DelCorro2013}, and Open IE-4.x~\cite{mausam2016open}) to extract relation triples from the raw text\footnote{The original text is produced from the WMT 2011 News Crawl data, at http://www.statmt.org/lm-benchmark/.}. These extractors have their expertise, so to ensure the diversity of extraction results. If a triple is obtained simultaneously by the three extractors, we should believe that the triple is correct and add it to our corpus.

All the extractors are based on the dependency parsing trees, so the extracted relational phrase may be a combination product. That is, the distance between the adjacent words in the phrase may be distant in the original sentence sequence. Moreover, the adjacency order of the words in a triple may be different from that in the sentence. For example, from the sentence ``He thought the current would take him out, then he could bring help to rescue me", we can get a triple (the current, would take out, him). 
Thus, we define some word order constraints: The arguments in a triple (Argument1, Relation, Argument2) are ordered. All words in Argument1 must appear before all words in Relation. All the words contained in Argument2 must appear after the Relation words. The order of words appearing in Relation must be consistent with them appearing in the original sentence sequence. In addition, each relational word must have appeared in the original sentence and is not the modified word or the added word. These constraints is to make a model easy to use sequence tagging.

We randomly sample 100 triples from the dataset to test the accuracy. The accuracy performance is up to 0.95. The sampling result verifies the validity of the above operating. In addition, it represents that the extraction noises caused by syntactic analysis errors can also be well filtered out.

In order to build a high-quality corpus for model training, we are committed to the high-accuracy of extractions at the expense of the recall. The constructed dataset is imperfect, but it still has acceptable scalability. The experimental sections prove the effectiveness of the dataset.

\subsection{Automatically Sequence Tagging}

Tagging is to assign a special label to each word in a sentence~\cite{zheng2017}. We use the BIOES annotation (Begin, Inside, Outside, End, Single) that indicates the position of the token in an argument or the relational phrase. It has been reported that this annotation is more expressive than others such as BIO~\cite{Ratinov2009CoNLL}.
The words appearing outside the scope of a relation triple will be labeled as ``O''.
As for a relation triple, the arguments and relational phrase could span several tokens within a sentence respectively. Thus, the arguments and relation-phrase need to be division and tagged independently.

In the general tagging scheme~\cite{Stanovsky2018Supervised}, only one label will be assigned to each word in sentences. However, A sentence may contain multiple, overlapping triples. An argument can have multiple labels when it belongs to different triples. As a result, the normal tagging program cannot handle the case where a relation is involved in multiple triples, or an argument is related to multiple relations.

To cope with such multiplicity, we design an overlap-aware tagging scheme that can assign multiple labels for each word. Each triple will correspond to its unique tag sequence.
Figure~\ref{fig:one} is an example that a sentence is tagged by our scheme. 
Giving an argument pair, there can only be one relation between the pair in a sentence~\footnote{The coordination of verbs in a sentence should be considered as one complete relational phrase.}. Therefore, after the argument tags being pre-identified, the sequence of relational tags is uniquely determined.

When using models to predict, we extract the candidate argument pairs in advance~\footnote{Candidate arguments recognition is easy to do through many existing methods, so we can add it to the pre-processing process. We use the methods in the ClausIE~\cite{DelCorro2013} and Open IE-4.x~\cite{mausam2016open} to pre-identify the arguments in sentences.}, then transform them into the argument embedding as model inputs to identify the relationship between them.

\subsection{From Tag Sequence to Extracted Results}

From the tag sequence  \emph{Tags Sequence1} in Figure~\ref{fig:one}, ``The America" and ``Trump" can be combined into a triple whose relation is ``President". Because the relation role of ``The America" is ``1" and ``Trump" is ``2", the final result is (The America, President, Trump). The same applies to (Trump, will visit, the Apple), (the Apple, founded by, Steven Paul Jobs).

\begin{figure*}
	\centering
	\includegraphics[width=0.99\textwidth]{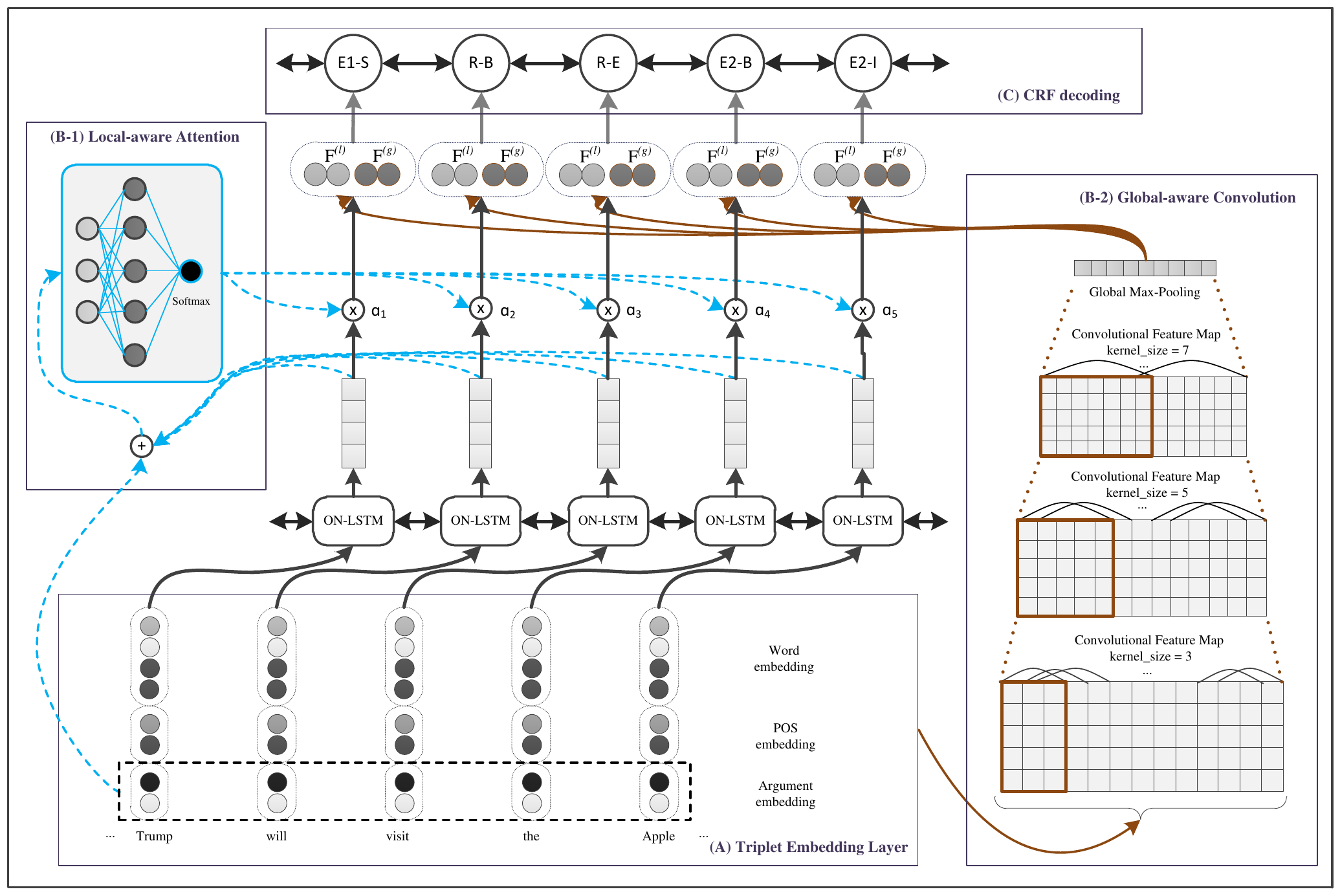}
	\caption{An illustration of our model.}
	\label{fig:two}
\end{figure*}

\section{Methodology }
\label{section-four}

We provide a detailed description of our model (HNN4ORT). It is illustrated in Figure~\ref{fig:two}. Our model has the following superiorities:
\begin{itemize}
\item[1] Employ the \textbf{Ordered Neurons LSTM} (ON-LSTM) to learn temporal semantics, while capturing potential syntactic information in natural language.
\item[2] Design \textbf{Dual Aware Mechanism}, including Local-aware Attention and Global-aware Convolution. It emphasizes focusing on local tagging but with a sentence-level semantics as a global perspective.
\end{itemize}

\subsection{ON-LSTM Network}
\label{subsection-4-1}

Natural language can be defined as sequence. However, the underlying structure of language is not strictly sequential. It should be hierarchically structured that be  determined by a set of rules or syntax~\cite{sandra2014morphological}. 

This syntactic information is critical to the relation extraction task. There are strict semantic associations and formal constraints that the head argument is the agent of the relation and the tail argument is the object of the relation. 
Integrating syntax-semantics into a neural network can encode better representations of natural language sentences. 

Recurrent neural network (RNN), and in particular it's variant  the Long short-term memory network (LSTM), is good at grasping the temporal semantics of a sequence. They are useful for all kinds of tagging applications~\cite{raganato2017neural, zheng2017,zhou2017word}. However, RNNs  explicitly impose a chain structure on a sentence.  The chain structure is contrary to the potential hierarchical structure of language, resulting in
that RNN (including LSTM) models may be difficult to effectively learn potential syntactic information~\cite{shen2018ordered}.

Therefore, we employ a new RNN unit, ordered neurons LSTM (ON-LSTM)~\cite{shen2018ordered}. It enables model to performing tree-like syntactic structure composition operations without destroying its sequence form. For a given sentence \(S=\{{x}_{1}, {x}_{2}, \ldots, {x}_{t}\}\), the ON-LSTM returns the representation \(\mathbf{H}=\{ \mathbf{h}_{1}, \mathbf{h}_{2}, \ldots , \mathbf{h}_{t} \}\) about the sequence $S$. 

The ON-LSTM uses a special inductive bias for the standard LSTM. This inductive bias promotes differentiation of the life cycle of information stored inside each neuron: high-ranking neurons will store long-term information, while low-ranking neurons will store short-term information. 

The ON-LSTM uses an architecture similar to the standard LSTM. 
An LSTM unit is composed of a cell memory and three multiplicative gates (i.e. input gate, forget gate, and output gate) which control the proportions of information to forget and to pass on to the next time step~\cite{hochreiter1997long,ma2016}. The forget gate $f_{t} $ and input gate $i_{t}$ in ON-LSTM are identical with them in LSTM, where they are used to control the erasing and writing operation on cell states $c_{t}$, as,

\begin{equation}
\centering
\mathbf{f}_{t}  = \sigma \left( \mathbf{W}_{f} \mathbf{x}_{t} + \mathbf{U}_{f} \mathbf{h}_{t - 1} + b_{f} \right)
\end{equation}
\begin{equation}
\centering
\mathbf{i}_{t}  = \sigma \left( \mathbf{W}_{i} \mathbf{x}_{t} + \mathbf{U}_{i} \mathbf{h}_{t - 1} + b_{i} \right).
\end{equation}

As for the cell memory $c_{t}$, it may be difficult  for a standard LSTM to discern the hierarchy information, since the gates in the LSTM act independently on each neuron. Therefore, the ON-LSTM enforces the order where the cell in each neuron should be updated.
The updated information in cells is allocated by the activation function $cumax()$. The contribution of the function is to produce the vectors of \textit{master input gate} $\tilde{i}_t$ and \textit{master forget gate} $\tilde{f}_t$ ensuring that when a given neuron is updated, all of the neurons that follow it in the ordering are also updated.

\begin{equation}
\centering
\hat{\mathbf{c}}_t  = \tanh \left( \mathbf{W}_{c} \mathbf{x}_{t} + \mathbf{U}_{c} \mathbf{h}_{t - 1} + b_{c} \right)
\end{equation}
\begin{equation}
\centering
\tilde{\mathbf{f}}_t  = cumax( \mathbf{W}_{\tilde{f}} \mathbf{x}_{t} + \mathbf{U}_{\tilde{f}} \mathbf{h}_{t - 1} + b_{\tilde{f}})
\end{equation}
\begin{equation}
\centering
\tilde{\mathbf{i}}_t  = cumax( \mathbf{W}_{\tilde{i}} \mathbf{x}_{t} + \mathbf{U}_{\tilde{i}} \mathbf{h}_{t - 1} + b_{\tilde{i}})
\end{equation}
\begin{equation}
\centering
\mathbf{\omega}_t  = \tilde{\mathbf{f}}_t \circ \tilde{\mathbf{i}}_t
\end{equation}
\begin{equation}
\centering
\mathbf{c}_t  = \mathbf{\omega}_t \circ \left( \mathbf{f}_{t} \circ \mathbf{c}_{t - 1} + \mathbf{i}_{t} \circ \hat{\mathbf{c}}_t \right) + \left( \tilde{ \mathbf{f}}_t - \mathbf{\omega}_t \right)\circ \mathbf{c}_{t - 1} + \left(\tilde{\mathbf{i}}_t - \mathbf{\omega}_t \right)\circ \hat{\mathbf{c}}_{t}
\end{equation}

According to the output gate $o_{t}$ and cell memory $c_{t}$, the output of the hidden state $\mathbf{h}_{t}$ is

\begin{equation}
\centering
\mathbf{o}_{t}  = \sigma \left( \mathbf{W}_{o} \mathbf{x}_{t} + \mathbf{U}_{o} \mathbf{h}_{t - 1} + b_{o} \right)
\end{equation}
\begin{equation}
\centering
\mathbf{h}_{t}  = \mathbf{o}_{t} \circ \tanh \left( \mathbf{c}_{t} \right).
\end{equation}

In the above formulas, $\mathbf{W}$ and $\mathbf{U}$ are the trainable parameter matrixes, $b$ is the bias. $\sigma$ is the logistic sigmoid function, $\circ$ denotes the Hadamard product.

\subsection{Local-aware Attention}
\label{subsection-4-2}

The RNN-based encoder generates the equal-level representation for each token of the entire sentence regardless of its contexts. However, a sentence may present an argument from various aspects, and various relations only focus on fractional aspects of the text.

Given the target arguments, not all of words/phrases in its text description are useful to model a specific fact. Some of them may be important for predicting relation, but may be useless for other relations. Empirically, relational words tend to appear in the neighborhoods of arguments.
Therefore, it is necessary for a model to pay more attention to the relevant parts of a sentence, according to different target relation. We propose a Local-aware Attention Network according to the attention mechanism~\cite{bahdanau2014neural,xu2017knowledge,huang2019knowledge}, as shown in Figure~\ref{fig:two}(B-1).

After being given a candidate argument pair $a$, the attention weight for each token $x_{t}$ of the sentence is defined as $\alpha_{t}$, which is

\begin{equation}
\alpha_{t}=\frac{\exp \left(\mathbf{V}_{t}\right)}{\sum_{i=1}^{T} \exp \left(\mathbf{V}_{i}\right)}
\end{equation}

\begin{equation}
\mathbf{V}_{t}(a)= \tanh \left(\mathbf{W}_{a} \mathbf{h}_{t}+\mathbf{U}_{a} \mathbf{a}\right).
\end{equation}

In the above formula, $\mathbf{W}_{a}$, $\mathbf{U}_{a}$ are parameters matrices. The $\mathbf{a}$ is the embeddings of the candidate argument pair. It is the key information that tells the model what the expected destination is, so that the same sentence can get different attention states for different predicting relations. The $\mathbf{h}_{t}$ is obtained by the bidirectional ON-LSTM network. The ON-LSTM processes the sequence $S$ in both directions and encodes each token $x_{t}$ into a fixed-size vector representation $\mathbf{H}=[\mathbf{h}_{1}, \mathbf{h}_{2}, \ldots, \mathbf{h}_{t}]$, by being calculated as
\begin{equation}
\mathbf{h}_{t}=\textit{ON-LSTM}_{fw}(\overrightarrow{\mathbf{h}_{t-1}},\mathbf{c}_{t}) \oplus \textit{ON-LSTM}_{bw}(\overleftarrow{\mathbf{h}_{t+1}},\mathbf{c}_{t}),
\end{equation}
where $\overrightarrow{\mathbf{h}_{t-1}}$ is the hidden representation of the forward ON-LSTM ($\textit{ON-LSTM}_{fw}$) at position $t-1$, and $\overleftarrow{\mathbf{h}_{t+1}}$ are the hidden representation of the backward ON-LSTM ($\textit{ON-LSTM}_{bw}$) at position $t+1$, $\oplus$ denotes concatenation operation.

We apply the attention weight $\alpha_{t}$ to $\mathbf{h}_{t}$, resulting in an updated representation $\mathbf{H}^{(a)}$, as

\begin{equation}
\mathbf{H}^{(a)}= \{\alpha_{1} \mathbf{h}_{1}, \alpha_{2} \mathbf{h}_{2}, \ldots, \alpha_{t} \mathbf{h}_{t} \}.
\end{equation}

\subsection{Global-aware Convolution}
\label{subsection-4-3}

Normal sequence labeling models cannot fully encode sentence-level information but encode word-level context information as assigning each word a tag to indicate the boundary of adjacent segments~\cite{Zhuo2016Segment, AAAI1714776}. However, we believe that the Local-aware Attention mechanism can play a better role only when it is provided with a comprehensive sentence-level context.
Therefore, in this section, we propose a Global-aware Convolution network. It and the Local-aware Attention network complement each other and form the Dual Aware Mechanism.

We use the stacked Convolutional neural network to learn global information from a whole sentence since it owns powerful spatial perception ability~\cite{collobert2011natural,shen2017deep}.
In the process of recognizing a piece of text, the human brain does not recognize the entire text at the same time, but firstly perceives each local feature in the text, and then performs a comprehensive operation at a higher level to obtain global information. To simulate this learning process of the human brain, we design the Global-aware Convolution network, as shown in Figure~\ref{fig:two}(B-2).

Fragment-level feature maps can be derived from a set of convolutions that operate on the embeddings of a sentence $S$. Convolution is an operation between a vector of weights and the embeddings. The weight matrix is regarded as the filter for the convolution~\cite{Zeng2014}. We apply a set of convolutional filters $\mathbf{W}_{l} $ and bias terms ${b}_{l}$ to the sentence as per equation~\ref{euq14}, to learn a representation.
\begin{equation}
\mathbf{h}_{t}^{l}=\textit{RELU}\left(\mathbf{W}_{l} \cdot \left[\mathbf{x}_{t}, \ldots, \mathbf{x}_{t+l}\right]+b_{l} \right),\; l = 2i+1 \;\;  \; i=1,2,...
\label{euq14}
\end{equation}
where $\mathbf{h}_{t}^{l}$ is used to represent vectors at time $t$, which includes all phrases of length $l$ started with $x_{t}$.
$i$ indicates the id of layers.

By stacking layers of convolutions of increasing filter width, we can expand the size of the effective input width to cover the most of length of a sequence.
Each iteration takes as input the result of the upper convolution layer. The kernel size of each current layer is 2 more than that of the upper layer.

After convoluting, a global max-pooling operation is applied that stores only the highest activation of each filter. It is used for feature compression and extracts the main features. 
So far, the final output vector with a fixed length can be obtained as sentence-level global features.

\subsection{Hybrid Framework}

Figure~\ref{fig:two} illustrates the framework of our model. The main components of the framework are the Triplet Embedding Layer, Dual Aware Encoder, and Tag Decoder.

We adopt triplet embeddings to map each token to a vector, as shown in Figure~\ref{fig:two}(A).
Given a sentence $S$ as a sequence of tokens (i.e., words of the sentence), each token $x_{t}$ is represented by 
a set of embeddings $(\mathbf{x}_{t}^{(w)}, \mathbf{x}_{t}^{(a)}, \mathbf{x}_{t}^{(p)})$, where $\mathbf{x}_{t}^{(w)}$ represents the word embedding, $\mathbf{x}_{t}^{(a)}$ denotes the embedding of candidate argument pair, and $\mathbf{x}_{t}^{(p)}$ shows the Part-of-speech (POS) embedding. We consider the POS information into our model since it plays an important role in the relation extraction processing. 

Next,  the token representation vectors $\mathbf{S}=\{\mathbf{x}_{1}, \mathbf{x}_{2}, \ldots, \mathbf{x}_{t}\}$ are sent to the Dual Aware Encoder. One side, these embeddings are fed into the module of Local-aware Attention, as shown in Figure~\ref{fig:two}(B-1), outputting the features $\mathbf{F}^{(l)}$.
Meanwhile, after giving input $\mathbf{S}$, we execute the deep convolution module (B-2) to output global-aware features $\mathbf{F}^{(g)}$.
Therefore, the encoder returns a representation $\mathbf{F} = \left[ \mathbf{F}^{(l)}; \mathbf{F}^{(g)} \right]$.

Subsequently, the upper output $\mathbf{F}$ is fed into the Tag Decoder to jointly yield the final predictions for each token, as shown in Figure~\ref{fig:two}(C). 
We take advantage of the sentence-level tag information by a Conditional random fields (CRF) decoding layer.
 
Notably, there are strong dependencies across output labels. It is beneficial to consider the correlations between labels in neighborhoods. The CRF can efficiently use past and future tags to predict the current tag. Therefore, it is a common way to model label sequence jointly using a CRF layer~\cite{lample2016, ma2016,yang2018design}.

In detail, for an output label sequence \(y= \left ( y_{1}, y_{2}, \ldots, y_{t}\right)\), we define its score as
\begin{equation}
\label{equ:crf1} 
s\left ( S, y \right )= \sum_{i=0}^{T}\mathbf{A}_{y_{i},y_{i+1}}+\sum_{i=1}^{T}\mathbf{P}_{i,y_{i}},
\end{equation}
where \(\mathbf{A}\) is a matrix of transition scores such that \(\mathbf{A}_{i,j}\) represents the score of a transition from the tag \(i\) to tag \(j\). \(y_{0}\) and \(y_{n}\) that separately means the start and the end symbol of a sentence. We regard \(\mathbf{P}\) as the matrix of scores outputted by the upper layer. \(\mathbf{P}_{i,j}\) corresponds to the score of the \(j^{th}\) tag of the \(i^{th}\) word in a sentence. 

We predict the output sequence that gets the maximum score given by
\begin{equation}
\label{equ:crf2} 
y^{*} =\underset{\tilde{y}\in Y_{S}}{\mathrm{argmax}}s\left ( S,\tilde{y} \right ),
\end{equation}
where \(Y_{S}\) represents all possible tag sequences including those that do not obey the BIOES format constraints.

\section{Experiments and Discussions}
\label{section-five}

In this section, we present the experiments in detail. We evaluate various models with Precision (P), Recall (R) and F-measure (F1). 

\subsection{Experimental Setting}

\textbf{Testing set}. To satisfy the openness and effectiveness of the experiments, we gather four testing sets from the previously published works. They should be close to nature and independent of the training set. Table 1 presents the details of the four datasets.

Firstly, the \textbf{Reverb dataset} is obtained from~\cite{Fader2011} which consists of 500 sentences with manually labeled 1,765 extractions. The sentences are obtained from Yahoo. Next, the \textbf{Wikipedia dataset} includes 200 random sentences extracted from Wikipedia. And we collect 605 extractions manual labeled by Del Corro~\cite{DelCorro2013}. Then, the \textbf{NYT dataset} contains 578 triples extracted from 200 random sentences in the New York Times collection. It is also created by Del Corro et al.
In addition, Stanovsky and Dagan~\cite{Stanovsky2016Creating} present the \textbf{OIE2016 dataset} which is an opened benchmark for ORE. It contains 10,359 tuples over 3200 sentences.

\begin{table*}
	\caption{The datasets for test in this work.}
	\label{one-table}
	\centering
	\begin{tabular}{lcccc}
		\toprule
		{Id}& {Dataset}&  {Source}& {Sent.} & {Triples} \\	\midrule
		{1 }& {Reverb dataset} 	& {Yahoo}	&{500} & {1,765} \\
		{2}& {Wikipedia dataset} 	& {Wikipedia}	&{200} &{605} \\
		{3}& {NYT dataset} 	& {New York Times}	&{200}  &{578} \\
		{4}& {OIE2016 dataset} 	& {QA-SRL annotations}	&{3200} & {10,359}  \\	\bottomrule
	\end{tabular}
	\label{tab2}
\end{table*}

\noindent
\textbf{Hyperparameters}. We implement the neural network by using the Keras library\footnote{https://github.com/keras-team/keras}. 
The training set and validation set contain 395,715 and 81,986 records, respectively. 
The batch size is fixed to 256. We use early stopping~\cite{graves2013} based on performance on the validation set. The number of LSTM units is 200 and the number of feature maps for each convolutional filter is 200. Parameter optimization is performed with Adam optimizer~\cite{kingma2014}. The initial learning rate is 0.001, and it should be reduced by a factor of 0.1 if no improvement of the loss function is seen for some epochs. Besides, to mitigate over-fitting, we apply the dropout method~\cite{srivastava2014} to regularize models.
We use three types of embeddings as inputs. Word embedding is pre-trained by word2vec~\cite{mikolov2013distributed} on the corpora. Its dimension is 300. Part-of-speech (POS) embedding is obtained by using the TreeTagger~\cite{Schmid1994Probabilistic} which is widely adopted to annotate the POS category, containing 59 different tags. It is 59 dimensions one-hot vectors. Besides, we represent the tag information of arguments as argument embedding, by using 10 dimensions one-hot vectors.

\subsection{ Experimental Results }

\begin{figure}
	\centering
	\includegraphics[width=0.7\textwidth]{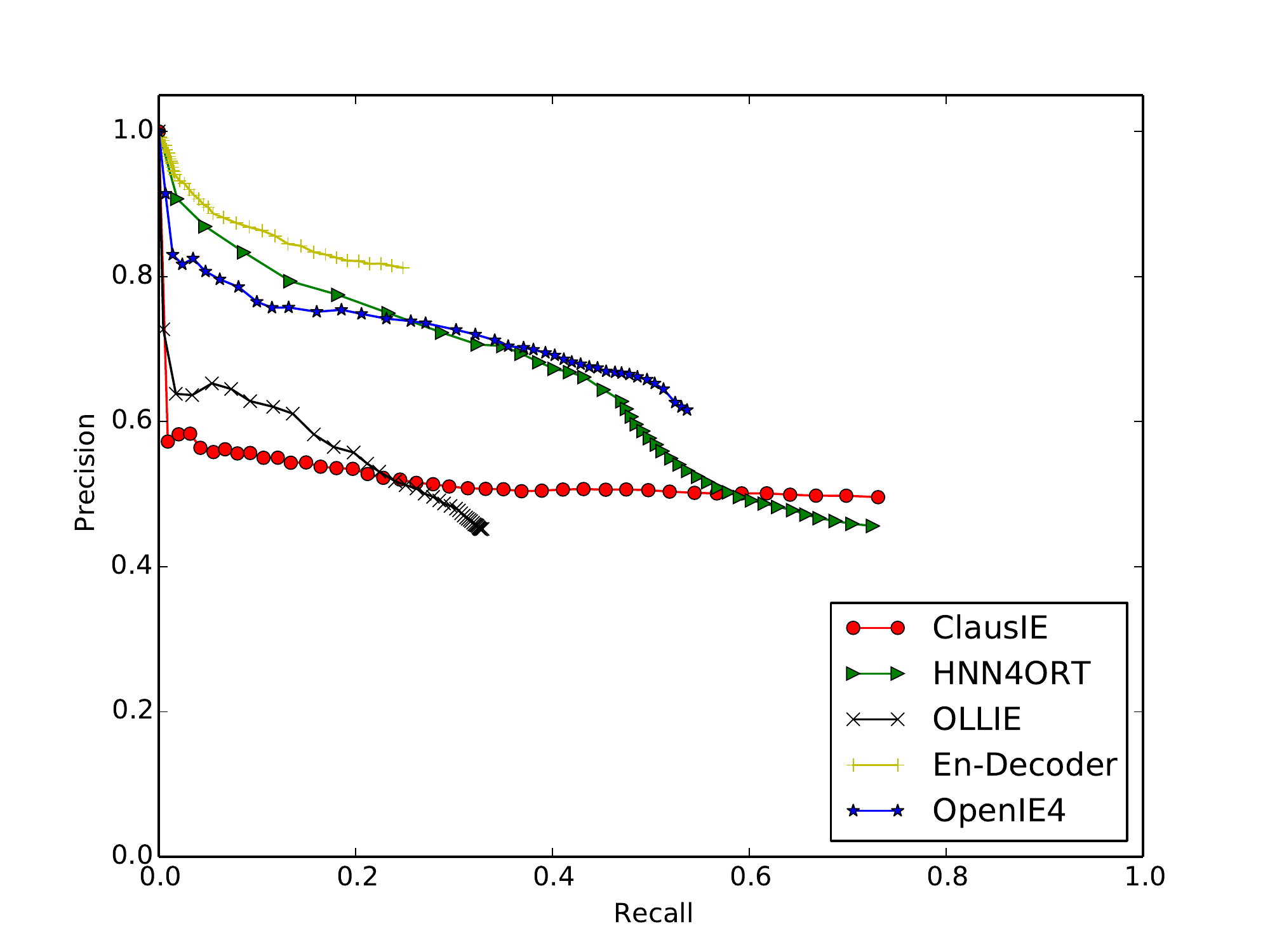}
	\caption{The Precision-recall curves of different ORE systems on the OIE2016 dataset. The model En-Decoder comes from  ~\cite{leicui}.}
	\label{fig:five}
\end{figure}

We report the results~\footnote{In order to avoid the distortion of argument recognition errors to the final performance, we only recognize the correctness of the relational phrases in a triple.}\footnote{As for a relation extracted by Reverb, OLLIE, ClausIE, and Open IE-4.x, only when its confidence is greater than 0.5 can it be judged that the relations are correct.} of various models to work on the Reverb dataset, Wikipedia dataset, and NYT dataset, as shown in Table~\ref{two-table} and Figure~\ref{fig:five}. We can draw the following conclusions:

(1) We discuss various popular neural structures and do lots of experiments to adapt to the open relation tagging task. It is valuable to understand the effectiveness of different neural architectures for the ORE task and to help readers reproduce experiments. These methods include the unidirectional LSTM network (\textit{UniLSTM})~\cite{ma2016}; the bidirectional LSTM network with a Softmax classifier (\textit{LSTM}), or with  
a CRF classifier (\textit{LSTM-CRF})~\cite{huang2015bidirectional, ma2016, lample2016, Chiu2016}; the CNN network with a Softmax classifier (\textit{CNN}), or with  
a CRF classifier (\textit{CNN-CRF})~\cite{collobert2011natural, shen2017deep}; the combination of the LSTM and CNN (\textit{LSTM-CNN-CRF})~\cite{zhai2017,feng2018language} and so on.

Our model HNN4ORT outperforms all the above methods according to F1. It shows the effectiveness of its hybrid neural network architecture. Remarkably, previous models (such as \textit{LSTM-CNN-CRF}~\cite{zhai2017,feng2018language}) have improved performances by integrating CNN and LSTM. They usually simply concatenate the feature vectors learned from the CNN and LSTM at each time-step.  However, we propose the Dual Aware Mechanism which achieves better performances.

In addition, we analyze the effects of various networks. Compared to unidirectional LSTM (\textit{UniLSTM}), bidirectional LSTM (\textit{LSTM}) is obviously superior to UniLSTM in F1 on average, since it can capture richer temporal semantic information. LSTM is better than CNN in recall and F1, however, CNN takes better precision. In addition, from the comparison between LSTM and LSTM-CRF, ON-LSTM and ON-LSTM-CRF, CNN and CNN-CRF, we can get a unanimous conclusion that the CRF layer can greatly improve model performances than a Softmax layer.

\begin{table}
	\caption{The predicted results of different models on the Reverb dataset, Wikipedia dataset, and NYT dataset.  The bolds indicate the best value when our model HNN4ORT compares to  other sequence tagging models. And comparing with the conventional ORE extractors, we highlight the best value with the underline.}
	\centering
	\label{two-table} 
	\begin{tabular*}{1.0\textwidth}{p{2.2cm}|p{0.55cm}p{0.55cm}p{0.7cm}|p{0.55cm}p{0.55cm}p{0.7cm}|p{0.55cm}p{0.55cm}p{0.7cm}|p{0.55cm}p{0.55cm}p{0.7cm}}\toprule 
		\small Model&\multicolumn{3}{c|}{ \small Wikipedia dataset}&\multicolumn{3}{c|}{ \small NYT dataset} &\multicolumn{3}{c|}{ \small Reverb dataset}&\multicolumn{3}{c}{ \small Average}\\\hline 
		\small 	&P&R&F&P&R&F&P&R&F&P&R&F\\\midrule
		\small CNN	&\bf 0.886&	0.527&	0.661&	\bf 0.931&	0.493&	0.645&\bf	0.915&	0.504&	0.650&\bf	0.912&	0.506&	0.651\\
		UniLSTM &	0.559&0.463	&0.506&	0.607&	0.481&	0.537&	0.632&	0.530&	0.576&	0.612&	0.506&	0.554\\
		\small LSTM &	0.766&	0.683&	0.722&	0.807	&0.637&	0.712&	0.784&	0.716&	0.748&	0.784&	0.693&	0.736\\
		\small ON-LSTM &	0.808 &0.714 &0.758&0.812 & 0.658 & 0.726&	0.816 & 0.728 & 0.769&	0.812 & 0.708 & 0.756\\
		\small CNN-CRF&	0.878&	0.574&	0.694&	0.854&	0.578&	0.689&	0.892&	0.568&	0.694&	0.882&	0.571&	0.693\\
		\small LSTM-CRF&	0.804&	0.734&	0.768	&0.781	&0.654&	0.712&	0.810&	0.743&	0.775&	0.804&	0.724&	0.762\\
		\small LSTM-CNN-CRF&	0.821&	0.729&	0.772&	0.837&\bf	0.702&	\bf 0.764&	0.831&	0.760&	0.794&	0.830&	0.743&	0.784\\\midrule
		\small ON-LSTM-CRF&	0.858&	0.719&	0.782&	0.825 & 0.663 &	0.735 &	0.849 &	0.731 &	0.786 &	0.846 &	0.715&	0.775\\
		\small  +Local-Aware  &	0.830&	0.749&	0.787&	0.818&	0.690&	0.749&	0.830&	0.760&	0.794&	0.828&	0.744&	0.784\\
		\small +Global-Aware &	0.869&	0.744&	0.801&	0.838&	0.656&	0.736&	0.867&	0.748&	0.803&	0.862&	0.729&	0.790\\
		\small  \bf HNN4ORT &0.841&\underline{\bf 0.759}&\underline{\bf 0.798}&0.825&\underline{0.678}&\underline{ 0.745}&0.859&\underline{\bf 0.778}&\underline{\bf 0.817}&0.849&\underline{\bf 0.754}&\underline{\bf0.799} \\
		\midrule
		\small Reverb&	0.770&	0.210&	0.330&	0.557&	0.144&	0.228&	0.595&	0.133&	0.217&	0.641&	0.162&	0.259\\
		\small {OLLIE}&	\underline{0.994}	&0.279&	0.436&	\underline{0.986}&	0.249&	0.398&	\underline{0.975}&	0.198&	0.329&	\underline{0.985}&	0.242&	0.389\\
		\small {ClausIE}	&0.795&	0.526&	0.633&	0.656	&0.481&	0.555&	0.953&	0.585&	0.725&	0.801&	0.531&	0.638\\
		\small {Open IE-4.x}&	0.766&	0.340&	0.471&	0.801&	0.341&	0.478&	0.810&	0.312&	0.451&	0.792&	0.331	&0.467\\\bottomrule
	\end{tabular*}
	
\end{table}

(2) We compare neural sequence models with conventional methods. When the three models used to construct corpus (OLLIE, ClausIE, and Open IE-4.x) being regarded as baselines, our model HNN4ORT achieves better recall and F1. Especially, our model achieves a 16.1\% improvement in F1 over the best baseline model ClausIE.

In addition, many of the neural sequence learning methods outperform conventional syntax-based methods. The conventional methods may be of high accuracy, but they can't learn rich enough patterns resulting in a lower recall. In addition, patterns is hard and inflexible, however, the syntax structures of sentences are ever-changing. The neural network-based models can learn deep sentence semantics and syntactic information, to achieve better precision and recall. 

(3) We verify the effectiveness of each module of our model HNN4ORT. 
First, we test the capability of the ON-LSTM. To our knowledge, it is the first time to be used for sequence tagging. The experimental results show that the ON-LSTM is better than LSTM about 2 \% in F1 on average. This shows that the potential syntactic information captured by ON-LSTM is very useful for open relation annotation.

Second, we use the ON-LSTM with the CRF classifier (\textit{ON-LSTM-CRF}) as the basic model, to test the performances of Dual Aware Mechanism.
After the Local-aware Attention network is integrated into the ON-LSTM-CRF network (\textit{+Local-Aware}), the effect is increased by about 1 \%. After the Global-aware Convolution network is integrated into the ON-LSTM-CRF network (\textit{+Global-Aware}), the effect is increased by about 1.5 \%. This proves that both modules are valid. 

Furthermore, when the dual modules are adopted at the same time, the model increases by more than 2 \% on the average of F1. It shows that the Dual Aware Mechanism is highly reasonable that global sentence-level information and local focusing information complement each other for handling open relation extraction.

\begin{figure}
	\centering
	\includegraphics[width=0.7\textwidth]{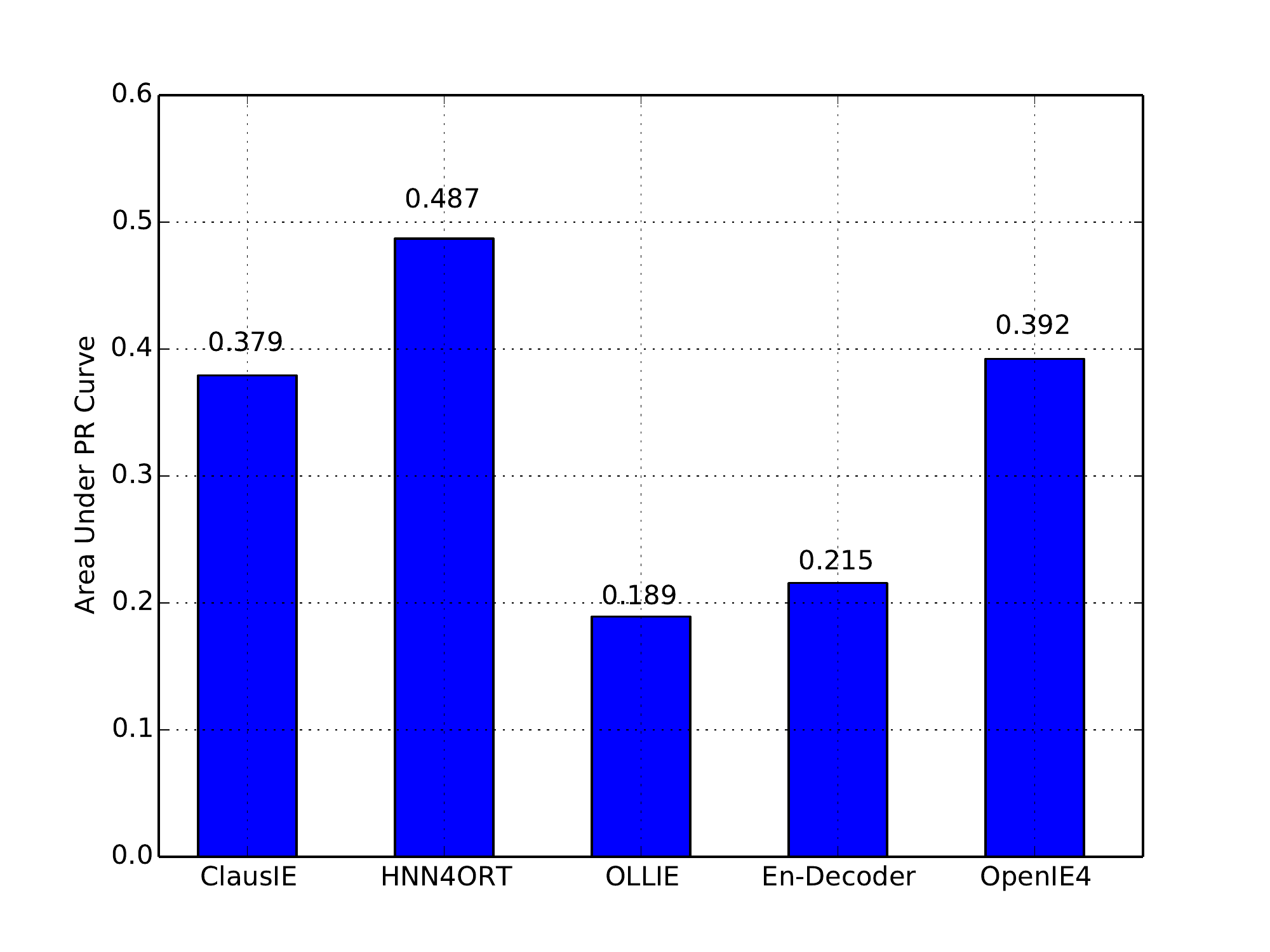}
	\caption{The Area under precision-recall curve (AUC) shown in Figure~\ref{fig:five}.}
	\label{fig:six}
\end{figure}

(4) We use the OIE2016 dataset to evaluate the precision and recall of different systems~\footnote{The model proposed in~\cite{Stanovsky2018Supervised} isn't shown here, because it used the OIE2016 dataset as a training set. We have evaluated it in Table~\ref{two-table}. Its model structure is the same as the BiLSTM network with a Softmax output layer (LSTM, i.e. the 4th line).}. The precision-recall curves are shown in Figure~\ref{fig:five}, and the Area under precision-recall curve (AUC) for each system is shown in Figure~\ref{fig:six}. It is observed that our model HNN4ORT has a better precision than the three conventional models in the most recall range. The HNN4ORT is learned from the bootstrapped outputs of the extractions of the three conventional systems, while the AUC score is better than theirs. It shows that the HNN4ORT has fixed generalization ability after training on the training set. Although the precision of the neural model En-Decoder~\cite{leicui} is better than that of the HNN4ORT, the recall of it has been maintained in a lower range than that of the HNN4ORT. In addition, the HNN4ORT achieves the best AUC score of 0.487, which is significantly better than other systems. In particular, the AUC score of the HNN4ORT is two times more than that of the En-Decoder model.

(5) To sum up, the performances of our model HNN4ORT on the four datasets are stable and superior. It indicates that the model has good robustness and scalability.

\section{Performance Analysis and Discussions}
\label{section-six}

\subsection{Evaluation of Training Corpus}

According to whether the three extractors that are used to construct the training set can correctly identify each test instance, we classify the test instances into four parts: all three extractors identify correctly (ATE), existing two extractors identify correctly (ETE), only one extractor identify correctly (OOE), no one extractor identify correctly (NOE). 

\begin{figure}
	\centering
	\includegraphics[width=0.58\textwidth]{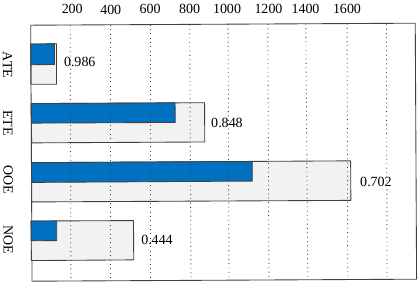}
	\caption{The performance of the model HNN4ORT in each part of the testing sets. Here, the black bar indicates the number of correct identifications of the model. And the gray bar represents the total number of this part.}
	\label{fig:three}
\end{figure}

As shown in Figure~\ref{fig:three}, the model HNN4ORT identifies these instances in the ATE with an accuracy close to 1. It implies that the model can learn the training set data features well.  In addition, from the perspective of a single extractor, the model HNN4ORT can acquire the extracting ability of this single extractor after training on our training set. The correct results extracted by this single extractor, in addition to appearing in the ATE, will also appear in the ETE and OOE. These instances outside the ATE own the cognate regularity (data distribution) with those in the ATE. Therefore, the HNN4ORT has a certain recognition in three parts (ATE, EYTE, and OOE). The accuracy is highest on the ATE, follows by it on the ETE, and worse on the OOE (but still above 0.7).

In particular, the model can produce certain results in the NOE where such kind of instances may be little or barely appear in the training set. It shows that the model has strong generalization ability and can obtain more powerful recognition capability beyond the any single extractor of the three extraction tools.
In addition, as shown in Table~\ref{two-table}, the model has achieved remarkable performances in various testing sets, although these testing sets and training set come from different sources. Based on the above phenomena, it can be concluded that the quality of training corpus should be acceptable.

Other than this, to maintain the high quality of the training set, we only select triples from the ATE. Although the ATE occupies a small proportion of the output of the extractors, the instances in the training set are considerable and are extracted from a large-scale text. This ensures that the training set has good diversification and passable quality.

\subsection{Analysis of  Embeddings}
We evaluate the effect of various embeddings, as shown in Table~\ref{three-table}. When the model HNN4ORT only uses word embeddings, the results are worse than that of the model with word embeddings and POS embeddings. We certificate that the POS features play a great role in promoting model performances and improve the F1 by 4.1\%. In addition, when the input embeddings are adjusted along with model training, the effect is better and the F1 is increased by 1.4\%.

\begin{table*}
	\caption{\label{three-table} The results of evaluating the influence of embeddings used on the model HNN4ORT. The results are average on the Reverb dataset, Wikipedia dataset, and NYT dataset.}
	\begin{center}
		\begin{tabular}{lccc}
			\toprule
			Model &  P &  R &  F\\
			\midrule
			(word embeddings)	&	0.783	&	0.708	&	0.744 \\
			(word, POS embeddings)	&0.822	&	0.751	&	0.785 \\
			(word, POS embeddings)$^{train}$		&0.849	&	0.754	&	0.799 \\
			\bottomrule
		\end{tabular}
	\end{center}
	
\end{table*}

\subsection{Analysis of Overlapping Triples}
As shown in Table~\ref{five-table}, The statistical results show that overlapping triples account for a large proportion in various datasets, all of which are above 30\%. By using our overlap-aware tagging scheme, models can identify multiple, over-lapping triples, so that they have a better recall capacity. We perform the model HNN4ORT to identify overlapping triples, with good results on all three datasets.

\begin{table}
	\caption{\label{five-table} The performances of the model HNN4ORT on the sub-dataset which contains only overlapping triples of each dataset.  The second column shows the proportion of overlapping triples in the total.}
	\begin{center}
		\begin{tabular}{l|c|ccc}
			\toprule  
			Datasets & Proportion & P	&  R	& F	 \\
			\midrule 
			Wikipedia &	38.7\% & 0.849	&0.650	&0.736
			\\
			NYT	& 31.1\% & 0.779	&0.606	&0.681	
			\\
			Reverb	&44.3\% & 0.794&	0.649 	&0.714	
			\\
			\bottomrule
		\end{tabular}
	\end{center}
	
\end{table}

\subsection{Error Analysis}

\begin{figure}
	\centering
	\includegraphics[width=0.68\textwidth]{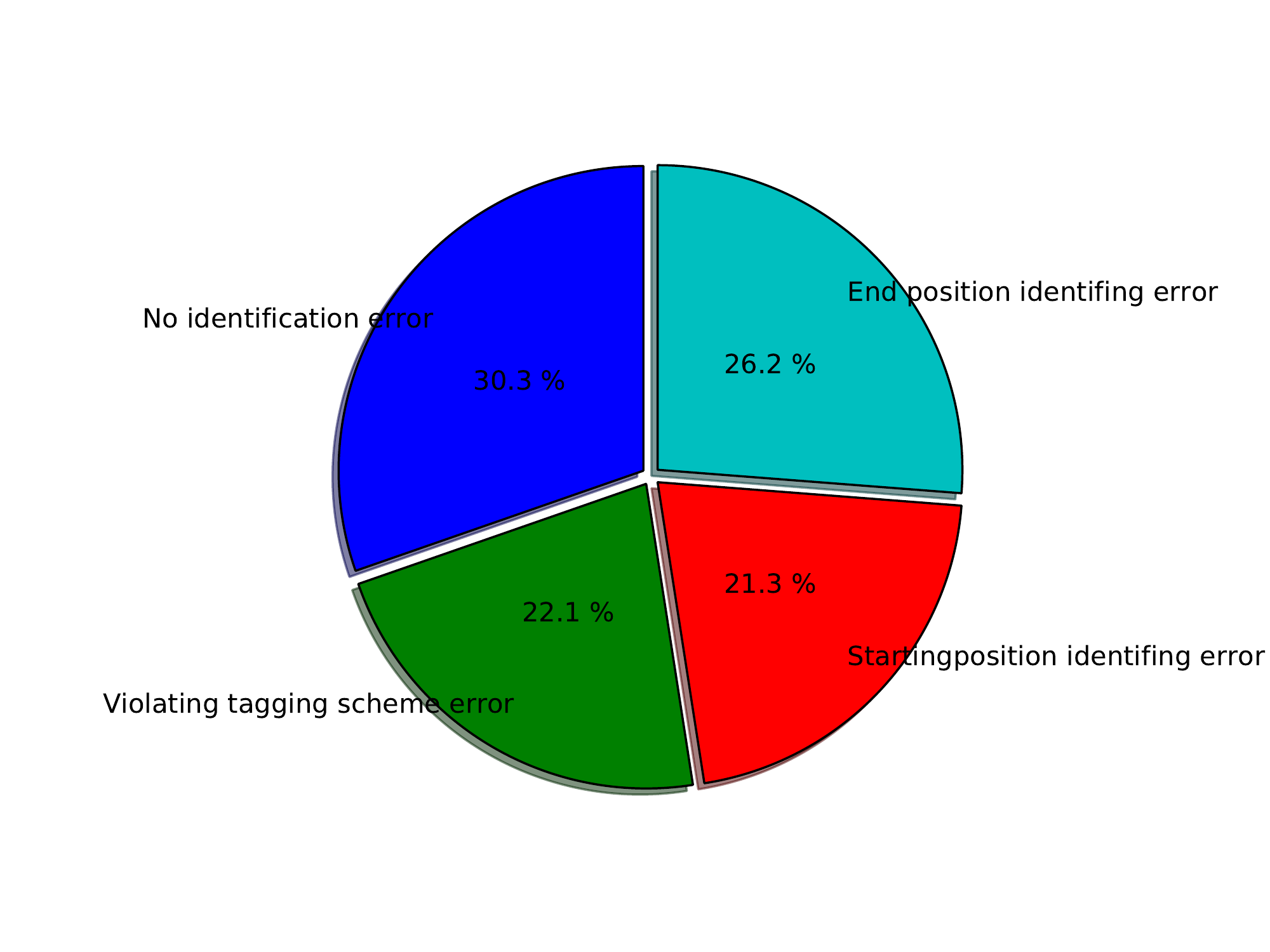}
	\caption{Error Results Analysis of the model HNN4ORT. There are four major types of errors.}
	\label{fig:four}
\end{figure}

To find out the factors that affect the performance of ORE tagging models, we analyze the tagging errors of the model HNN4ORT as Figure~\ref{fig:four} shown. There are four major types of errors. The 30.3\% relations are missed by the model. And 22.1\% of the extractions are abandoned because their corresponding tag sequences violate the tagging scheme. The two types of errors mainly limit the increase in recall. In addition, the model may wrongly determine the start or end position of a relational phrase. As a result, the relation will be recognized as falseness. Such phenomena affect model precision.

\section{Conclusions}
\label{section-seven}

Open relation extraction is an important NLP task. The effect of conventional methods is not satisfactory. Therefore, we are committed to using advanced deep learning models to solve this task. In this paper, we construct a training set automatically and design a neural sequence tagging model to extract open relations. Taking the ON-LSTM and Dual Aware Mechanism as the essence, we propose a hybrid neural sequence tagging model (NHH4ORT). The experimental results show the effectiveness of our method. Compared with conventional extractors or other neural models, our approach achieves state-of-the-art performances on multiple testing sets. We also analyze the quality of the automatically constructed training set, and conclude that it should be highly acceptable. In future work, we will consider to study a more efficient annotation scheme and use it to deal with n-ary relation tuples.

\bibliographystyle{ACM-Reference-Format-Journals}
\bibliography{ACMTIST.bib}

\end{document}